\documentclass[letterpaper]{article} 
\usepackage{aaai2026}  
\nocopyright

\usepackage{times}  
\usepackage{helvet}  
\usepackage{courier}  
\usepackage[hyphens]{url}  
\usepackage{graphicx} 
\usepackage{natbib}  
\usepackage{caption} 
\usepackage{algorithm}
\usepackage{algorithmic}
\usepackage{booktabs}
\usepackage{amsmath}
\usepackage[utf8]{inputenc}
\DeclareUnicodeCharacter{2212}{-}

\usepackage{newfloat}
\usepackage{listings}
\DeclareCaptionStyle{ruled}{labelfont=normalfont,labelsep=colon,strut=off} 
\floatstyle{ruled}
\newfloat{listing}{tb}{lst}{}
\floatname{listing}{Listing}
\newcommand{\daggernote}{\textsuperscript{\textdagger}}

\title{Equilibrium Dynamics and Mitigation of Gender Bias
in Synthetically Generated Data}
\author{
    Ashish Kattamuri\daggernote\textsuperscript{\rm 1},
    Arpita Vats\daggernote\textsuperscript{\rm 2},
    Harshwardhan Fartale\textsuperscript{\rm 3}, \\
    Rahul Raja\daggernote\textsuperscript{\rm 2},
    Akshata Kishore Moharir\daggernote\textsuperscript{\rm 3},
    Ishita Prasad\daggernote\textsuperscript{\rm 4}
}
\affiliations{
    \textsuperscript{\rm 1}Proofpoint \thanks{This work does not relate to the authors’ positions at Proofpoint, LinkedIn, Microsoft, or Meta.}
    \textsuperscript{\rm 2}LinkedIn \textsuperscript{\rm 3}Independent Research
    \textsuperscript{\rm 4}Microsoft
    \textsuperscript{\rm 5}Meta FAIR

}

\usepackage{bibentry}

\begin{document}

\maketitle

\begin{abstract}
Recursive prompting with large language models enables scalable synthetic dataset generation but introduces the risk of bias amplification. We investigate gender bias dynamics across three generations of recursive text generation using three complementary evaluation frameworks:
rule-based pattern matching, embedding based semantic similarity, and downstream task performance. Experiments with three initial bias levels (0.1, 0.3, 0.6) and four mitigation strategies reveal equilibrium dynamics rather than monotonic amplification. The low initial bias amplifies toward the model's inherent bias level (+ 36\%), whereas the high initial bias decays toward it (-26\%). Among mitigation methods, contrastive augmentation, which introduces gender-swapped variants, achieves significant downstream bias reduction (98.8\% for low initial bias and 91\% on average) despite producing higher embedding-based bias scores. This paradox demonstrates that semantic similarity metrics may diverge from behavioral fairness outcomes, highlighting the need for multidimensional evaluation in responsible synthetic data generation.

\end{abstract}

\section{Introduction}

Foundation models increasingly generate synthetic training data through iterative prompting and self-refinement. While this approach enables scalable dataset creation, the bias implications of recursive synthetic generation remain insufficiently examined. The self-instruct framework proposed by \citet{wang2022self} transformed instruction tuning by allowing language models to produce diverse and high-quality examples from minimal seed data. Building on this idea, recursive variants reuse model outputs as inputs for subsequent generations, offering the potential for unlimited dataset expansion but also raising questions about how bias propagates and evolves over time.

Bias in large language models has been widely documented across a range of linguistic and reasoning tasks. Prior studies have revealed systematic gender, racial, and occupational biases in model representations and outputs \citep{bolukbasi2016man,zhao2018gender,bender2021dangers}. Early work by \citet{zhao2018gender} demonstrated strong occupational stereotyping in coreference resolution through the WinoBias benchmark, while subsequent evaluations such as BBQ \citep{parrish2022bbq} extended bias assessment to question-answering, revealing persistent disparities across model scales and architectures. These findings underscore that even well-trained models internalize and reproduce societal stereotypes embedded in their training data.

Amplification of such biases during model usage has emerged as a critical concern. Empirical evidence indicates that repeated inference or self-conditioning can exacerbate existing imbalances. For example, \citet{zhao2017men} observed that models tend to magnify training-set biases when generating new examples. More recently, \citet{wang2024bias} showed that iterative text continuation amplifies bias by 15-30\% over multiple generations, suggesting that recursive or self-referential processes can compound representational skew.

Despite these insights, bias dynamics in synthetic data generation remain largely unexplored. The recursive generation of instructions or examples introduces feedback loops where a model effectively learns from its own outputs, potentially reinforcing or equilibrating biases over time. Understanding these recursive effects is essential as synthetic data increasingly substitutes or supplements human-curated datasets in model training.

Various mitigation techniques have been proposed to address bias propagation. Data augmentation through gender swapping \citep{zhao2018gender}, adversarial debiasing \citep{zhang2018mitigating}, and content filtering \citep{welbl2021challenges} have all shown promise in constrained settings. Among these, contrastive augmentation, which creates paired gender variants of the same prompt, is notable for its simplicity and conceptual alignment with balance-oriented generation. However, its behavior under recursive synthetic generation has not been systematically studied.

In this work, we examine gender bias dynamics across three recursive generations of synthetic instruction data, using Google's Gemma-2-2b-it model as a case study. We evaluate how initial seed bias influences amplification trajectories and compare four mitigation strategies, including contrastive augmentation. Our analysis employs both rule-based and embedding-based bias metrics, along with downstream behavioral evaluation. 

The results suggest that recursive generation does not lead to inevitable bias growth but instead exhibits equilibrium dynamics, where systems stabilize around a model-specific bias level regardless of initialization. Notably, contrastive augmentation achieves substantial downstream bias reduction (91\% on average) even when embedding-based bias appears higher. This divergence highlights the limitations of single-metric evaluations and underscores the need for multidimensional fairness assessment in responsible synthetic data generation.

\section{Methodology}
We conducted recursive text generation experiments using Google's Gemma-2-2b-it model with a temperature of 0.7 across three recursive generations. Each seed produced five child outputs per generation, yielding a progression of 50 seeds $\rightarrow$ 250 (Gen-1) $\rightarrow$ 1,250 (Gen-2) $\rightarrow$ 6,250 (Gen-3). 

Seed sets were created at three target bias levels (0.1, 0.3, and 0.6) by sampling from a curated list of occupations: 12 female-associated roles (e.g., nurse, secretary, teacher), 12 male-associated roles (e.g., engineer, CEO, developer), and 20 gender-neutral prompts. Each seed consisted of a topic-oriented instruction such as ``Describe the responsibilities of a nurse,'' allowing for controlled bias measurement while preserving realistic recursive generation dynamics. 

We compared four recursive generation strategies that differ in how gender-related information is introduced, balanced, or filtered. Table~\ref{tab:strategies} summarizes the setup and rationale for each condition.

\begin{table}[h]
\centering
\renewcommand{\arraystretch}{1.1}
\setlength{\tabcolsep}{4pt}
\small
\caption{Summary of recursive generation strategies. Each strategy represents a distinct approach to controlling gender information during recursive synthesis.}
\label{tab:strategies}
\begin{tabular}{@{}p{2.3cm}@{}p{6.4cm}@{}}
\toprule
\textbf{Strategy} & \textbf{Description} \\[4pt]
\midrule
\textit{Vanilla} & Standard recursive generation without modification; serves as the baseline condition. \\[4pt]
\textit{Contrastive} & Introduces gender-swapped augmentation, pairing each gendered prompt with its opposite variant (e.g., ``male nurse'' and ``female engineer''); balances gender representation. \\[4pt]
\textit{Filtered} & Removes instructions with a rule-based bias score above 0.4, suppressing strongly stereotyped examples while maintaining data diversity. \\[4pt]
\textit{Size-matched} & Adds neutral instructions to match the sample size of the contrastive condition, isolating content effects from dataset size. \\
\bottomrule
\end{tabular}
\end{table}

\subsection{Bias Measurement}

To comprehensively assess bias evolution, we employed three complementary evaluation frameworks capturing distinct dimensions of bias: explicit lexical patterns, implicit semantic associations, and downstream behavioral effects. This multi-level approach enables a deeper understanding of how bias manifests and propagates across recursive generations.

\textbf{Rule-Based Metric.}  
Explicit gender bias was measured through pattern-based analysis of stereotypical co-occurrences between gendered pronouns (he/she, his/her) and occupation terms. Following prior work \citep{zhao2018gender}, the bias rate was computed as the proportion of stereotypical associations among all gendered instructions, as defined in Equation~\ref{eq:rule}:
\begin{equation}
\text{Bias}_{\text{rule}} = 
\frac{\text{Count(stereotypical pairs)}}{\text{Total gendered instructions}}.
\label{eq:rule}
\end{equation}
This formulation captures overt lexical bias that reflects surface-level gender associations in the generated text. Although simple, it provides an interpretable baseline for observing explicit bias amplification trends.\\

\textbf{Embedding-Based Metric.}  
To capture more subtle semantic biases, we used the \texttt{all-MiniLM-L6-v2} sentence transformer to generate instruction embeddings and compared them with gender prototype vectors. The prototypes were computed as the mean embeddings of male-associated and female-associated seed instructions. For each instruction $x$, we calculated cosine similarity with both prototypes, and an instruction was labeled as biased if its similarity margin exceeded 0.35, as shown in Equation~\ref{eq:embed}:
{\small
\begin{equation}
\text{Bias}_{\text{embed}}(x) =
\begin{cases}
1, & \text{if } |\cos(x, v_{male}) - \cos(x, v_{female})| > 0.35 \\
0, & \text{otherwise.}
\end{cases}
\label{eq:embed}
\end{equation}
}

This metric captures implicit bias expressed through representational proximity rather than explicit lexical markers. It reflects how the model organizes gendered concepts in semantic space, even when gender-specific words are not explicitly mentioned.

\textbf{Downstream Evaluation.}  
To evaluate whether instruction-level bias affects model behavior, we trained logistic regression classifiers on instruction embeddings to predict gender associations. We report both classification accuracy and a bias score defined as the absolute difference in predicted probabilities between male and female classes, as described in Equation~\ref{eq:down}:
\begin{equation}
\text{Bias}_{\text{down}} = 
|p(\text{male}) - p(\text{female})|.
\label{eq:down}
\end{equation}
This downstream bias metric quantifies behavioral disparities arising from representational differences. High values of $\text{Bias}_{\text{down}}$ indicate that even subtle embedding-level imbalances can translate into observable behavioral effects, linking representational bias and surface-level lexical bias within a unified evaluation framework.

Finally, we analyze all three bias measures across recursive generations to compare how explicit, implicit, and behavioral bias evolve under different generation strategies.

\section{Results}

\subsection{Equilibrium Dynamics in Vanilla Condition}

\begin{figure*}[t]
\centering
\includegraphics[width=0.32\textwidth]{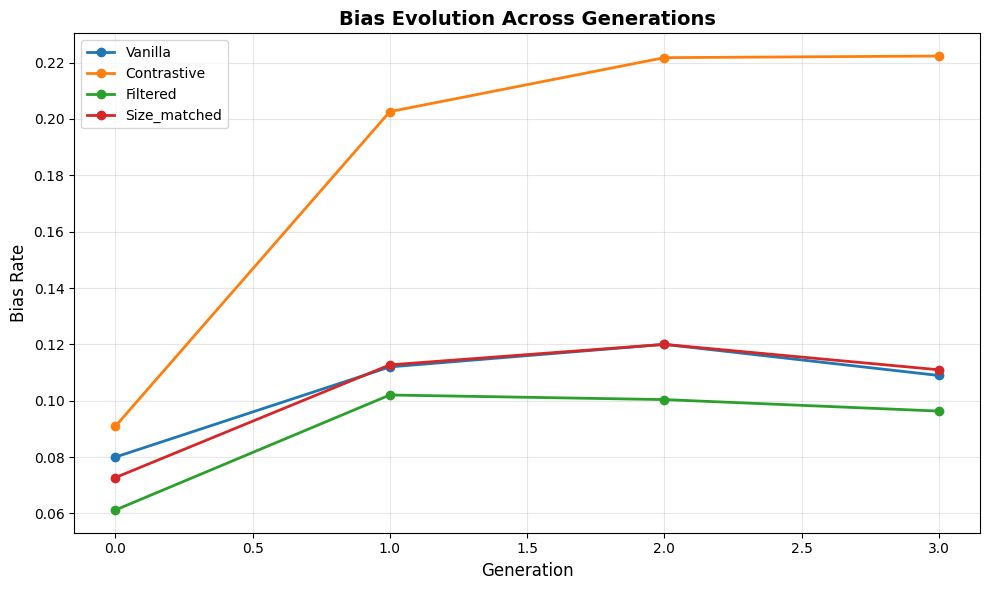}
\includegraphics[width=0.32\textwidth]{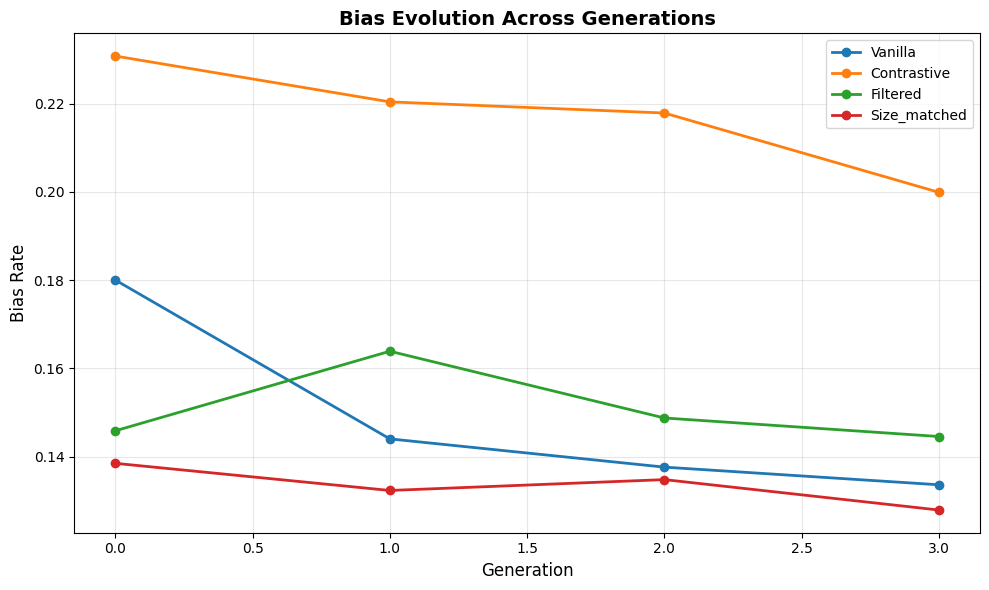}
\includegraphics[width=0.32\textwidth]{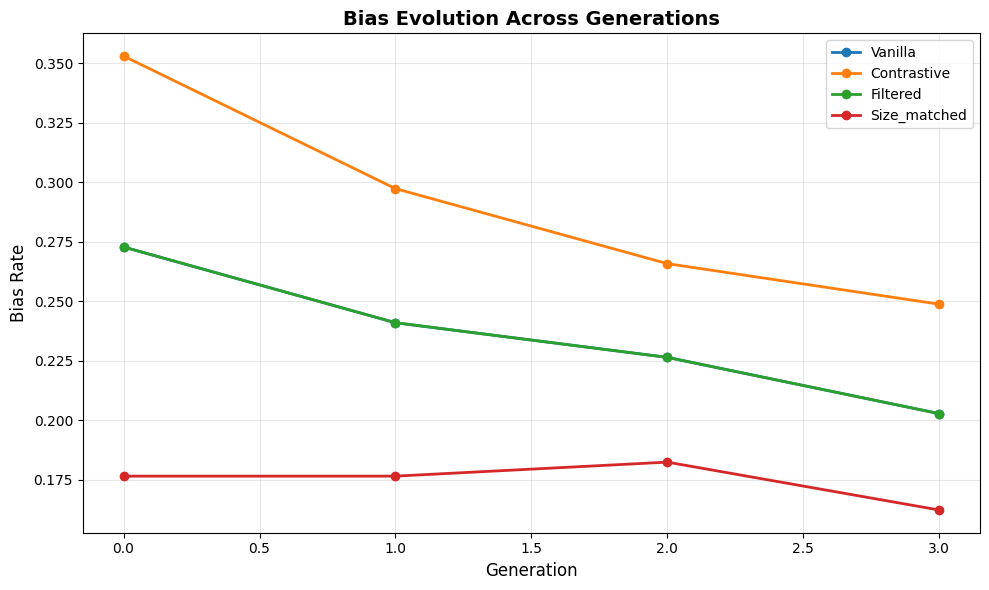}
\caption{Embedding bias evolution across three generations for initial bias levels 0.1 (left), 0.3 (center), and 0.6 (right). Vanilla (blue) demonstrates equilibrium dynamics, where low bias amplifies and high bias decays. Contrastive (orange) yields higher embedding bias but lower downstream bias.}
\label{fig:trajectories}
\end{figure*}

Embedding bias evolves in a non-monotonic manner across recursive generations. The vanilla generation condition demonstrates equilibrium dynamics, where systems converge toward a stable bias level over time rather than continuously amplifying or decaying. This behavior is illustrated in Figure~\ref{fig:trajectories}, which shows embedding bias trajectories across three initial bias levels (0.1, 0.3, 0.6) and three recursive generations.

Quantitative results in Table~\ref{tab:vanilla_results} confirm this equilibrium pattern. For low initial bias (0.1), embedding bias increased from 0.080 to 0.109 (+36\%), whereas for medium (0.3) and high (0.6) initial biases, it decreased by approximately 26\%. These changes indicate convergence toward a steady-state bias between 0.11 and 0.13, suggesting that the model possesses an inherent equilibrium bias level.

\begin{table}[t]
\centering
\small
\caption{Embedding and rule-based bias rates across generations under the vanilla condition. Systems converge toward equilibrium regardless of initial bias level.}
\label{tab:vanilla_results}
\begin{tabular}{@{}llcccc@{}}
\toprule
\textbf{Bias} & \textbf{Metric} & \textbf{Gen-0} & \textbf{Gen-1} & \textbf{Gen-2} & \textbf{Gen-3} \\
\midrule
0.1 & Embedding & 0.080 & 0.112 & 0.120 & 0.109 (+36\%) \\
    & Rule      & 0.200 & 0.167 & 0.267 & 0.342 (+71\%) \\
\midrule
0.3 & Embedding & 0.180 & 0.144 & 0.138 & 0.134 (-26\%) \\
    & Rule      & 0.467 & 0.560 & 0.557 & 0.579 (+24\%) \\
\midrule
0.6 & Embedding & 0.273 & 0.241 & 0.226 & 0.203 (-26\%) \\
    & Rule      & 0.542 & 0.545 & 0.542 & 0.535 (-1\%) \\
\bottomrule
\end{tabular}
\end{table}

Interestingly, rule-based bias followed a different trajectory. It showed monotonic growth for the low-bias condition (0.200 to 0.342, +71\%), a moderate increase for medium bias at +24\%, and near stability for high bias at negative 1\%. This divergence between rule-based and embedding-based measures suggests that they capture distinct dimensions of bias evolution, representing lexical and representational perspectives respectively. This reinforces the importance of multi-metric evaluation when analyzing recursive bias behavior.

\subsection{Mitigation Strategy Comparison}

Downstream bias, as measured by Equation~\ref{eq:down}, varied substantially across the four recursive generation strategies. The results summarized in Table~\ref{tab:downstream} reveal clear differences in mitigation effectiveness and show that the relationship between embedding-level bias (Equation~\ref{eq:embed}) and behavioral fairness is not always consistent.

\begin{table}[t]
\centering
\small
\caption{Downstream bias scores by strategy and initial bias level. Contrastive augmentation achieves a 91\% average reduction despite exhibiting higher embedding bias (Fig.~\ref{fig:trajectories}).}
\label{tab:downstream}
\begin{tabular}{@{}lccc|c@{}}
\toprule
\textbf{Strategy} & \textbf{Bias 0.1} & \textbf{Bias 0.3} & \textbf{Bias 0.6} & \textbf{Average} \\
\midrule
Vanilla & 0.424 & 0.140 & 0.057 & 0.207 \\
\textbf{Contrastive} & \textbf{0.005} & \textbf{0.009} & \textbf{0.039} & \textbf{0.018} \\
Filtered & 0.241 & 0.278 & 0.057 & 0.192 \\
Size-matched & 0.424 & 0.124 & 0.108 & 0.219 \\
\midrule
\textbf{Reduction (\%)} & \textbf{−98.8} & \textbf{−93.6} & \textbf{−31.6} & \textbf{−91.3} \\
\bottomrule
\end{tabular}
\end{table}

Among all strategies, contrastive augmentation achieved the most effective bias mitigation. Although it produced the highest embedding bias in Figure~\ref{fig:trajectories} (orange lines), its downstream bias was minimal. For low initial bias (0.1), the downstream score decreased from 0.424 in the vanilla setting to 0.005, corresponding to a 98.8\% reduction. Medium and high bias conditions showed similar improvements (93.6\% and 31.6\% reductions respectively), yielding a 91.3\% average reduction overall. This finding demonstrates that increased representational separation in embedding space does not necessarily translate to behavioral unfairness.

The filtered strategy displayed inconsistent results. It moderately improved fairness for low bias (−43\%), degraded it for medium bias (+99\%), and had negligible impact for high bias. These results suggest that filtering, while reducing explicit lexical bias, may also remove valid data and reduce sample diversity, leading to unstable mitigation outcomes.

The size-matched control performed slightly worse than the vanilla condition on average (+5.8\%), confirming that the improvement observed in the contrastive setting originates from content balancing rather than sample size effects.

\begin{figure}[t]
\centering
\includegraphics[width=0.48\textwidth]{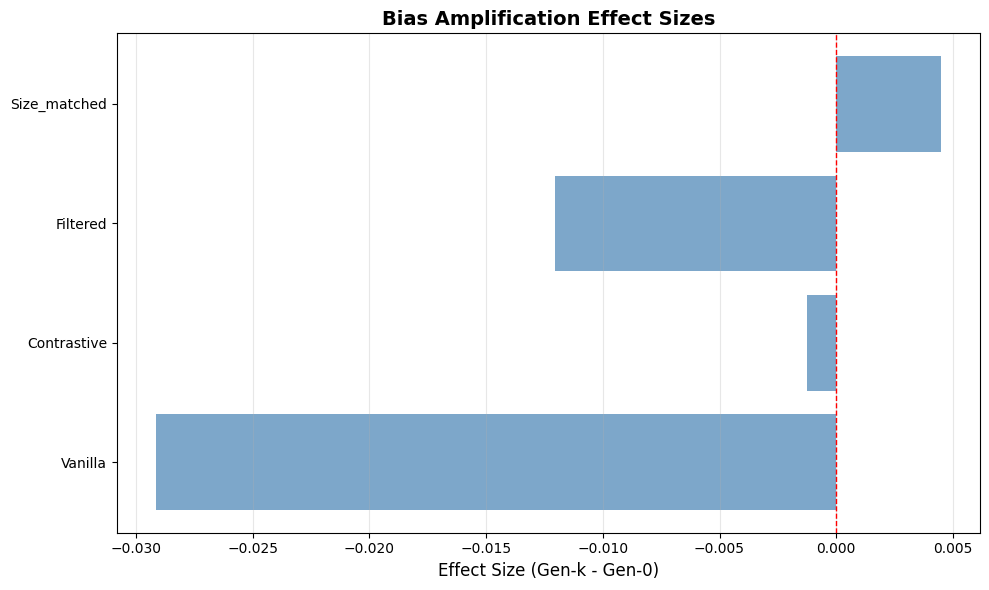}
\caption{Effect sizes (Gen-3 minus Gen-0) for embedding bias across strategies. Negative values indicate bias decay. Vanilla shows the strongest decay, while contrastive augmentation achieves the best downstream fairness.}
\label{fig:effect_sizes}
\end{figure}

Changes in embedding bias across generations are visualized in Figure~\ref{fig:effect_sizes}. The vanilla condition exhibits the strongest decay, while contrastive augmentation shows a small positive shift in embedding bias yet achieves the highest downstream fairness scores.

\begin{figure}[t]
\centering
\includegraphics[width=0.45\textwidth]{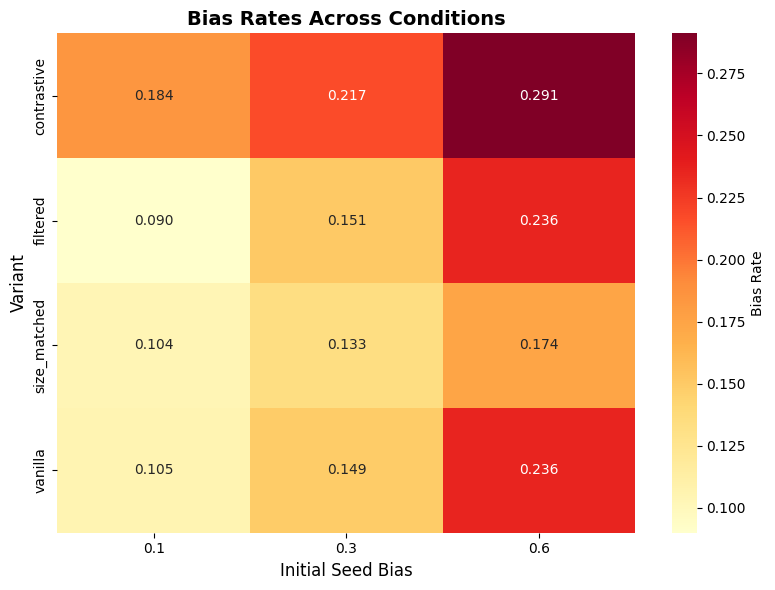}
\caption{Gen-3 embedding bias across strategies and initial bias levels. Darker shades represent higher embedding bias. Despite these higher values, contrastive augmentation achieves the lowest downstream bias.}
\label{fig:heatmap}
\end{figure}

Figure~\ref{fig:heatmap} provides a complementary view by visualizing Gen-3 embedding bias rates across all strategies and initial bias levels. Contrastive augmentation consistently yields higher embedding bias values than other strategies but simultaneously achieves the lowest downstream bias, reinforcing that embedding-level separation and behavioral fairness capture distinct dimensions of bias.

\subsection{Statistical Analysis}

Embedding bias distributions were compared across strategies through permutation testing with 5000 iterations. The analysis found no statistically significant differences between variants after false discovery rate (FDR) correction (contrastive versus vanilla: $p = 0.800$; filtered versus vanilla: $p = 0.809$; size-matched versus vanilla: $p = 0.394$). 

Although intrinsic differences in embedding bias were statistically indistinguishable, downstream bias results revealed large practical effects. The contrastive strategy achieved a 91\% reduction in downstream bias relative to the vanilla baseline, indicating that improvements in behavioral fairness can occur even when embedding-level changes are not statistically significant. This distinction emphasizes that statistical significance in intrinsic metrics does not necessarily correspond to practical significance in model behavior.

\section{Discussion}

Our findings provide three key insights with direct implications for responsible synthetic data generation.

\textbf{Equilibrium Dynamics Over Amplification.}  
Recursive generation does not lead to universal bias growth. Instead, the Gemma-2-2b-it model exhibits equilibrium dynamics, maintaining an intrinsic bias level around 0.11 to 0.13 as measured by the embedding metric. Seeds initialized below this level amplify toward it, while those above decay toward it. This behavior resembles regression to the mean in statistical systems. Effective mitigation approaches should therefore focus on shifting the equilibrium bias level itself rather than solely modifying the initial seed bias.

\textbf{The Contrastive Paradox.}  
Contrastive augmentation reveals an important paradox: higher embedding bias, reflecting stronger semantic polarization, coincides with substantially lower downstream bias and improved behavioral fairness. This occurs because gender-swapped augmentation produces two balanced semantic clusters, ensuring equal representation across genders. Embedding metrics capture representational separation, not fairness outcomes, whereas downstream bias reflects actual behavioral differences in model predictions. These results suggest that contrastive augmentation is effective precisely because it equalizes model outputs despite increased representational divergence.

\textbf{Multidimensional Bias Measurement.}  
The divergence among rule-based, embedding-based, and downstream bias measures highlights that bias is inherently multidimensional. Rule-based metrics capture explicit linguistic associations, embedding-based metrics quantify semantic clustering, and downstream evaluation assesses behavioral fairness. A comprehensive understanding of model bias requires integrating all three perspectives, since reliance on any single metric risks mischaracterizing mitigation outcomes.

\textbf{Limitations.}  
This study has several limitations. Computational constraints restricted the analysis to three recursive generations, and longer chains may reveal different convergence behaviors. The results are based on a single model, Gemma-2-2b-it, whose equilibrium level may not generalize to other architectures. Moreover, the binary gender framework used here does not capture non-binary or intersectional identities, which remain important directions for future work.

\section{Conclusion}

This study examined gender bias dynamics in recursive synthetic data generation and found equilibrium behavior rather than monotonic amplification. Low initial bias amplified toward the model’s inherent bias level (+36\%), while high initial bias decayed toward it (−26\%). Contrastive augmentation achieved a 91\% average reduction in downstream bias despite higher embedding bias, demonstrating that semantic clustering metrics can diverge from behavioral fairness outcomes. These findings indicate that effective mitigation must account for model-specific equilibrium dynamics and evaluate success through downstream task performance. As synthetic data generation becomes increasingly prevalent, understanding these equilibrium mechanisms is essential for designing responsible and bias-aware AI systems.

\bibliography{aaai2026}

\end{document}